\documentclass[conference]{IEEEtran}
\IEEEoverridecommandlockouts
\usepackage{cite}
\usepackage{amsmath,amssymb,amsfonts}
\usepackage{algorithmic}
\usepackage{graphicx}
\usepackage{textcomp}
\usepackage{xcolor}
\usepackage{xspace}
\usepackage{booktabs}
\usepackage{multirow}
\usepackage[table]{xcolor}
\usepackage{hyperref}

\newcommand{\name}[0]{GroupEnsemble}
\newcommand{\mc}[0]{MC-Dropout}

\newcommand{\gt}{\ensuremath >}
\newcommand{\ie}{i.e.\@\xspace}
\newcommand{\eg}{e.g.\@\xspace}

\newcommand{\vone}[1]{\textcolor{black}{#1}}
\newcommand{\vtwo}[1]{\textcolor{black}{#1}}
\newcommand{\vth}[1]{\textcolor{black}{#1}}
\newcommand{\katarina}[1]{\textcolor{black}{#1}}
\newcommand{\crv}[1]{\textcolor{black}{#1}}

\def\BibTeX{{\rm B\kern-.05em{\sc i\kern-.025em b}\kern-.08em
    T\kern-.1667em\lower.7ex\hbox{E}\kern-.125emX}}
\begin{document}

\title{GroupEnsemble: Efficient Uncertainty Estimation for DETR-based Object Detection
\thanks{
\textsuperscript{1} Mercedes-Benz AG, Germany

\textsuperscript{2} University of Stuttgart, Germany

\textsuperscript{3} Friedrich-Alexander University Erlangen-Nuremberg, Germany

\textsuperscript{\dag} Correspondence to yutong.yang@mercedes-benz.com

This work is funded by the Federal Ministry for Economic Affairs and Energy within the project "nxtAIM".}
}

\author{\IEEEauthorblockN{Yutong Yang\textsuperscript{1,2 \dag}}
\and
\IEEEauthorblockN{Katarina Popović\textsuperscript{1,3}}
\and
\IEEEauthorblockN{Julian Wiederer\textsuperscript{1,3}}
\and
\IEEEauthorblockN{Markus Braun\textsuperscript{1}}
\and
\IEEEauthorblockN{Vasileios Belagiannis\textsuperscript{3}}
\and
\IEEEauthorblockN{Bin Yang\textsuperscript{2}}
}
\maketitle

\begin{abstract}
Detection Transformer (DETR) and its variants show strong performance on object detection, a key task for autonomous systems.
However, a critical limitation of these models is that their confidence scores only reflect semantic uncertainty, failing to capture the equally important spatial uncertainty. 
This results in an incomplete assessment of the detection reliability.
On the other hand, Deep Ensembles can tackle this by providing high-quality spatial uncertainty estimates.
However, their immense memory consumption makes them impractical for real-world applications.
\vone{A cheaper alternative, Monte Carlo (MC) Dropout, suffers from high latency due to the need of multiple forward passes during inference to estimate uncertainty.}

\vtwo{To address these limitations, we introduce GroupEnsemble, an efficient and effective uncertainty estimation method for DETR-like models.}
\name{} simultaneously predicts multiple individual detection sets by feeding additional \vth{diverse} groups of object queries to the transformer decoder during inference.
Each query group is transformed by the shared decoder in isolation and predicts a complete detection set for the same input.
An attention mask is applied to the decoder to prevent inter-group query interactions, \vth{ensuring} each group detects independently \vth{to achieve} reliable ensemble-based uncertainty estimation.
\vtwo{By leveraging the decoder's inherent parallelism, \name{} efficiently estimates uncertainty in a single forward pass without sequential repetition.}
We validated our method under autonomous driving scenes and common daily scenes using the Cityscapes and COCO datasets, respectively.
\vtwo{The results show that a hybrid approach combining MC-Dropout and \name{} outperforms Deep Ensembles on several metrics at a fraction of the cost.
The code is available at \url{https://github.com/yutongy98/GroupEnsemble}.
}
\end{abstract}


\section{INTRODUCTION}
Object detection is a cornerstone of computer vision, essential in safety-critical domains like autonomous driving~\cite{b1,b2}. 
\vth{Conventional object detectors built upon convolutional architectures} have made significant progress, \vth{however they }often rely on hand-crafted components like non-maximum suppression (NMS) and anchor priors~\cite{b3,b31}. 
Recently, Detection Transformer (DETR)~\cite{b5} has emerged as a novel approach to object detection, reformulating the task as a set prediction problem without hand-crafted components. 

\begin{figure}[htbp]
\centerline{\includegraphics[width=0.9\columnwidth]{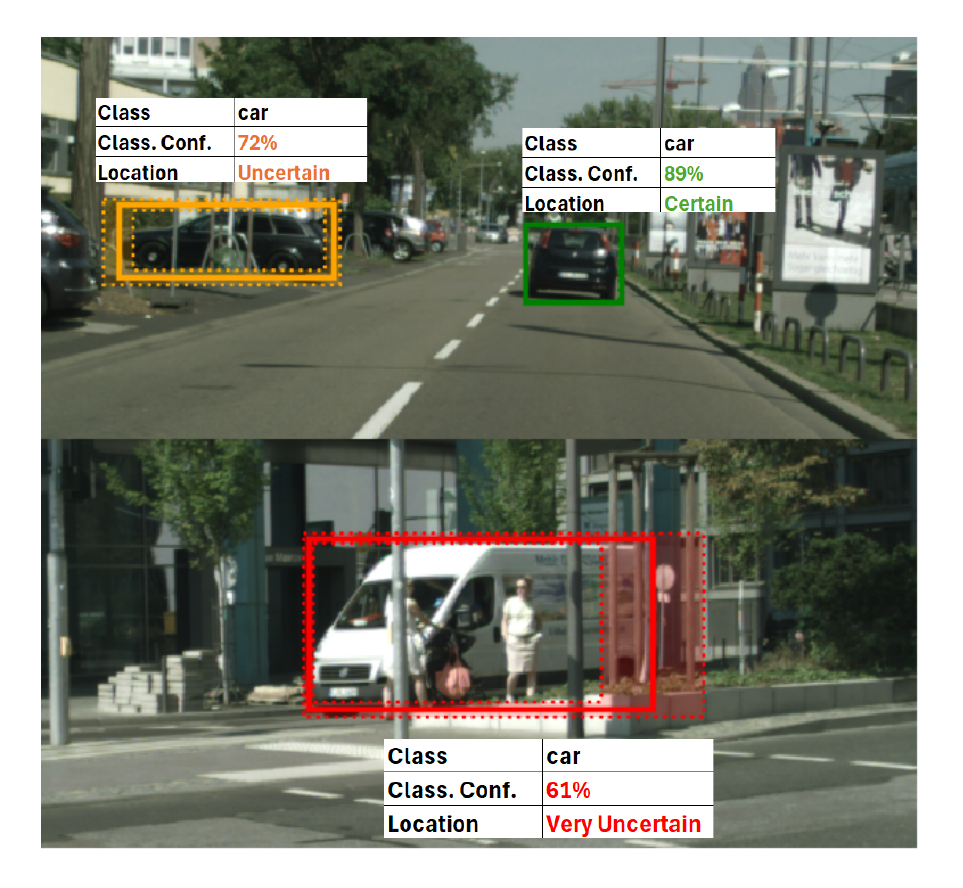}}
\vspace{-10pt}
\caption{
\textbf{Visualizations of semantic and spatial uncertainties \vth{ estimated by \name{}}.}
Semantic uncertainty is represented by the classification confidence (Class. Conf.), while spatial uncertainty is visualized using 95\% confidence intervals (dashed lines) around the mean (solid line). 
The color coding indicates the level of uncertainty: green for certain detections, yellow for medium uncertainty, and red for high uncertainty.
Factors such as occlusion or background clutter can lead to uncertainty in the location or size of the detected objects. 
\vth{By estimating both uncertainties, \name{} provides a comprehensive assessment of the reliability of the detections.}}
\label{fig:teaser}
\vspace{-6mm}
\end{figure}
\vth{However, beyond mere accuracy, reliable quantification of both semantic uncertainty (\ie, uncertainty regarding the object's class and its existence) and spatial uncertainty (\ie, uncertainty regarding the object's location and size) is paramount for object detection.
Such quantification are especially crucial in safety-critical applications where models must comprehensively recognize their limitations.}
\vth{Given an image, current object detectors predict a complete detection set containing the objects of interest.}
Each detection in the set consists of a class label with an associated confidence score and a bounding box.
While this confidence score can represent semantic uncertainty, the deterministic predicted bounding box fails to capture the spatial uncertainty.
A fundamental limitation arises from the fact that classification and localization are distinct tasks.
Consequently, a high-confidence semantic label prediction does not necessarily guarantee a similarly high level of certainty in the bounding box location.
Relying solely on the semantic uncertainty while neglecting spatial uncertainty can misrepresent the detection's overall uncertainty, potentially leading to serious consequences in safety-critical applications.
Therefore, it is essential to explicitly estimate spatial uncertainty in conjunction with semantic uncertainty to obtain a comprehensive assessment of detection reliability, as illustrated in Fig.~\ref{fig:teaser}.

For uncertainty estimation, Monte Carlo (MC) Dropout~\cite{b7} and Deep Ensembles~\cite{b8} are the two most popular methods~\cite{b6}. 
\vone{
Although MC-Dropout has been adapted for object detection to quantify spatial uncertainty, existing works have also reported its high latency issue, stemming from the need for multiple forward passes during inference~\cite{b9, b10,b29}.
Deep Ensembles share this high latency issue and additionally require considerable more memory due to the storage of multiple independent models. 
Such high computational cost is prohibitive for real-time applications like autonomous driving.}
Furthermore, to the best of our knowledge, there are no uncertainty estimation techniques specifically tailored for DETR-like detectors, \vth{as the current techniques} overlook their unique architectural properties for this purpose.

In this work, we introduce GroupEnsemble, an efficient and effective uncertainty estimation method tailored for DETR-like models. 
\name{ } bridges the gap between efficiency and uncertainty estimation accuracy for object detection.
\vth{Inspired by Group DETR~\cite{b18}, our method feeds additional diverse groups of object queries to the trained transformer decoder \textit{during inference}. }
Each query group is transformed by the shared decoder individually and predicts a complete detection set for the same input image.
To ensure independent detections across groups, an attention mask is applied to the decoder to prevent inter-group query interactions. 
In this way, \name{} predicts multiple independent and diverse detection sets, which is crucial for ensemble-based uncertainty estimation. 
Furthermore, by leveraging the decoder’s inherent parallelism, these multiple detection sets are  simultaneously obtained in \textit{a single pass} \vth{without sequential repetition.}

We evaluate \name{} across diverse scenarios, including the Cityscapes~\cite{b12} and Foggy Cityscapes~\cite{b14} datasets for autonomous driving scenes, and the COCO dataset~\cite{b13} for common daily scenes.
\vtwo{The results show that GroupEnsemble performs comparably to MC-Dropout.}
\vtwo{Moreover, a hybrid approach, MC-GroupEnsemble, which combines \name{} with MC-Dropout, outperforms strong Deep Ensembles on certain metrics with significantly lower computational cost.
}

The main contributions of this work are as follows:
\begin{itemize}
  \item We introduce GroupEnsemble, an efficient and effective uncertainty estimation method for DETR-based object detection.
  \item \vtwo{
    For the first time, \crv{we leverage DETR's inherent parallelism to achieve fast, single-pass uncertainty estimation for both GroupEnsemble and MC-Dropout.}
  }
  \item \vtwo{We show that MC-GroupEnsemble outperforms strong Deep Ensembles on various metrics at a significantly lower computational cost across three datasets.}
\end{itemize}

\section{RELATED WORK}
\subsection{DETR-based Object Detection}
DETR~\cite{b5} revolutionized object detection by introducing a transformer-based end-to-end set prediction framework.
Many follow-up works have proposed various DETR-like variants to improve the vanilla DETR. 
For instance, some variants modify the architecture, such as Conditional DETR~\cite{b15}, which represents implicit object queries as explicit 2D reference points for faster training convergence, and Deformable DETR~\cite{b16}, which leverages deformable attention for more efficient feature sampling.
Other works have focused on proposing universal training methods to accelerate DETR's training process without changing the model architecture.
For example, DN-DETR~\cite{b17} stabilizes training by adding an auxiliary denoising task, where the model learns to reconstruct original ground-truth (GT) boxes from noised versions. 
Similar to DN-DETR, Group DETR~\cite{b18} also feeds additional groups of object queries to the decoder for training augmentation. 
However, unlike DN-DETR, Group DETR randomly initializes these query groups as learnable embeddings and then supervises each group independently.
Notably, during inference, only one query group is used, while the remaining groups, despite their equivalently strong detection capabilities, are discarded. 

Our method builds upon Group DETR because it already provides multiple learned, high-quality groups of queries, which are essential for predicting multiple detection sets for uncertainty estimation.
Furthermore, because Group DETR is a universal training method applicable to almost any DETR-like model, our GroupEnsemble method inherits this flexibility.

\subsection{Uncertainty Estimation}
The goal of Uncertainty Estimation (UE) is to provide a quantified measure of the uncertainty or confidence associated with model prediction~\cite{b6}. 
A well-calibrated uncertainty estimate should accurately reflect the reliability of the corresponding prediction. 
In practice, this means that the uncertainty estimate should be high when the model is likely to make a wrong prediction and low when \vth{the prediction is correct.}
To estimate uncertainty, a leading family of methods called \textit{ensemble} produces multiple redundant yet diverse predictions for the same input~\cite{b6,b7,b8}, often as an approximation of Bayesian inference. 
The uncertainty among these predictions is then quantified through an aggregated measure such as variance or entropy~\cite{b22}. 
To this end, Monte Carlo (MC) Dropout~\cite{b7} and Deep Ensembles~\cite{b8} stand out as two of the most prominent and widely adopted methods~\cite{b6, b19}. 
MC-Dropout~\cite{b7} involves training a single stochastic network with dropout layers. 
At inference time, the dropout layers remain active, allowing multiple predictions to be produced by repeatedly feeding the same input to the stochastic model. 
Deep Ensembles~\cite{b8} , on the other hand, involve training multiple independent models with different random initializations.
Miller et al.~\cite{b9} was the first study which applies \mc{} to object detection for spatial uncertainty estimation.
Given an input image, it first uses \mc{} to obtain multiple individual detection sets and then applies Basic Sequential Algorithmic Scheme (BSAS)~\cite{b20} method to cluster these detections.
For each cluster, its final bounding box is the average of the bounding boxes within this cluster, and the corresponding spatial uncertainty is captured by the covariance matrix of those box coordinates.
\begin{figure*}[t!]
      \centering
      \includegraphics[width=\textwidth]{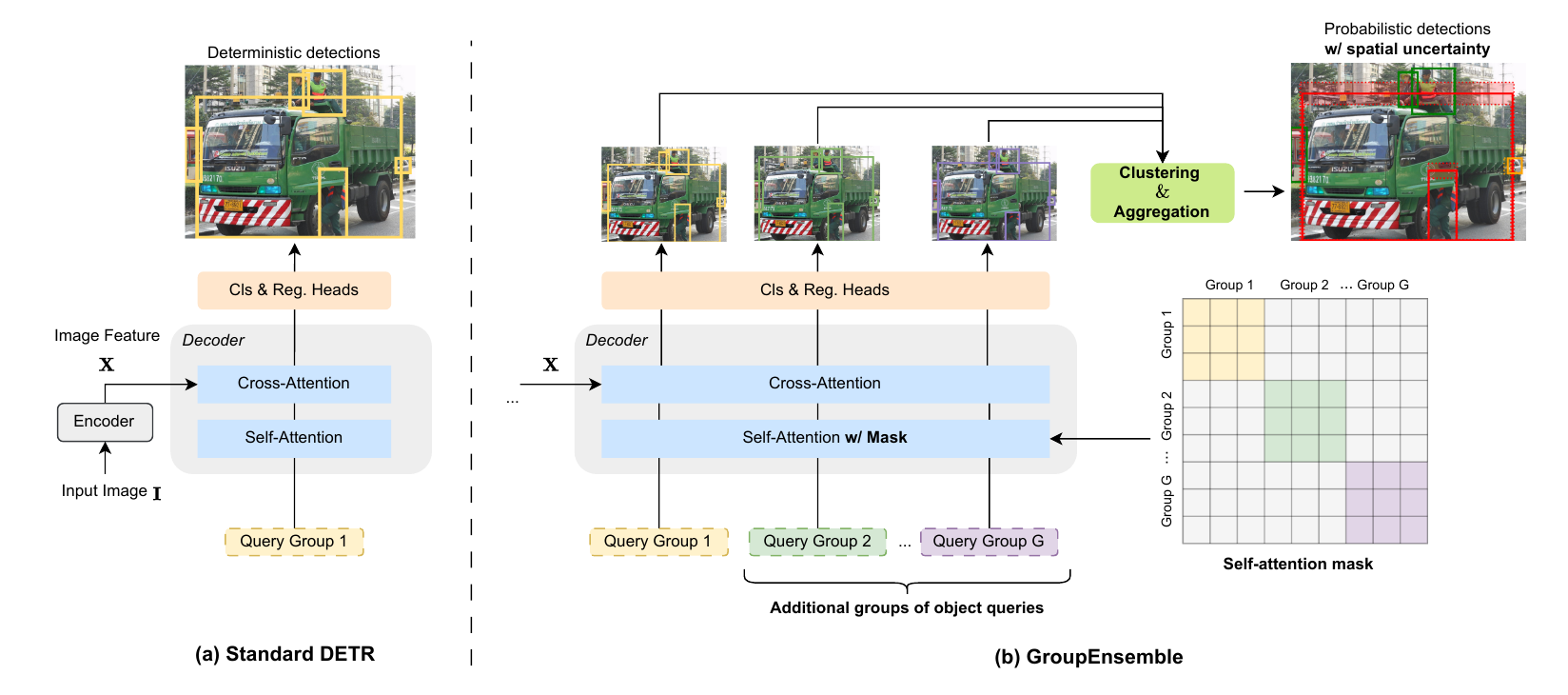}
      \vspace{-20pt}
      \caption{
        \textbf{Overview of \name{}.}
        For an input image, a standard DETR (a) outputs deterministic detections. 
        In contrast, our \name{} (b) feeds $G-1$ additional groups of object queries to the decoder and employs a self-attention mask (gray grids indicate blocked attentions) to block all inter-group query interactions.
        This enables the decoder to independently and simultaneously transform each query group, predicting multiple individual detection sets in a single pass. 
        These detections are then clustered and aggregated to produce final detections with semantic and spatial uncertainty estimates.
        Spatial uncertainty is depicted as dashed confidence intervals around the mean box.
        Larger intervals indicate higher uncertainty. \eg, for the person occluded by the truck door.
      }
      \vspace{-2mm}
      \label{fig:overview}
  \end{figure*}
Follow-up studies have also utilized MC-Dropout for uncertainty estimation in tasks such as 3D vehicle detection~\cite{b10} and \katarina{depth estimation~\cite{b30}} due to its adaptability and performance.

\vone{However, a major drawback of ensemble-based methods such as MC-Dropout is that their computational time scales linearly with the number of forward passes required~\cite{b29}. 
This high latency is often prohibitive for real-time applications like autonomous driving.}
\vone{This reveals a clear need for new approaches that can provide accurate uncertainty estimates without this significant computational cost.}
\vtwo{
To address this need in object detection, we introduce \name{}, the first method to leverage the unique properties of the DETR architecture for this goal.
}


\section{METHOD}
We first provide an overview of our GroupEnsemble (see Fig.~\ref{fig:overview}) and the underlying DETR model, followed by a detailed explanation of each key component.

\subsection{Overview}\label{AA}
DETR~\cite{b5} is composed of three key parts: an image encoder, a transformer decoder, and classification and regression heads.
Taking an image $\textbf{I}$ as input, the encoder outputs the image features $\textbf{X}$:
\begin{equation}
\text{Encoder}(\textbf{I})\to \textbf{X}.
\label{encoder}
\end{equation}

The transformer decoder takes as inputs the image features $\textbf{X}$ and $N$ learnable object queries denoted as $\textbf{Q} =\left\{ \textbf{q}_{1} \textbf{q}_{2}...\textbf{q}_{N}  \right\} $, and outputs the transformed queries $\tilde{\textbf{Q}}$,
\begin{equation}
\text{Decoder}(\textbf{X},\textbf{Q})\to \tilde{\textbf{Q}}.
\label{decoder}
\end{equation}

The transformer decoder utilizes two types of attention mechanisms: self-attention and cross-attention. 
Self-attention enables interactions between queries, while cross-attention allows the queries to probe image features $\textbf{X}$ and extract object-centric information for detection.

Finally, the classification and regression heads take as inputs the transformed queries $\tilde{\textbf{Q}}$, and output the detection set $\textbf{Y} =\left\{ \textbf{y}_{1} \textbf{y}_{2}...\textbf{y}_{N}  \right\}$:
\begin{equation}
\text{TaskHeads}(\tilde{\textbf{Q}})\to \textbf{Y}.
\label{head}
\end{equation}
Each detection $\textbf{y}=\left\{ \textbf{b}, k, c\right\}\in \textbf{Y}$ is described by its bounding box $\textbf{b}\in \mathbb{R}^{4}$, the predicted class $k\in \left\{ 1,2,...K \right\}$, and a confidence score $c\in \left[ 0,1 \right]$ about the class prediction, where $K$ is the total number of classes.

During training, DETR uses the Hungarian algorithm for bipartite matching~\cite{b24} between the detection set $\textbf{Y}$ and the set of ground-truth (GT) objects $\bar{\textbf{Y}}$.
A combined classification and regression loss, which we refer to as the Hungarian loss, is then calculated based on the matching results.
During inference, the learned object queries $\textbf{Q}$ probe image feature through the decoder, ultimately producing the detection set $\textbf{Y}$.

In addition to the standard DETR model, our \name{} incorporates three key components during inference:
\subsubsection{Additional Groups of Object Queries}
The key feature of \name{} is the use of multiple groups of object queries during inference, rather than relying solely on the single query group $\textbf{Q}$. 
In addition to $\textbf{Q}$, which contains $N$ queries, we introduce extra $G-1$ groups of object queries, each with $N$ queries.
In total, we have $G$ groups of object queries during inference, denoted as $\mathcal{Q}=\left\{ \textbf{Q}_{1},\textbf{Q}_{2},...,\textbf{Q}_{G} \right\}$.

\subsubsection{Attention Mask}
To ensure that each query group $\textbf{Q}_{g} \in \mathcal{Q}$ detects independently, we apply a carefully designed attention mask to the decoder, preventing any inter-group interactions while still allowing full intra-group interactions during self-attention. 
This allows each query group to be transformed independently and simultaneously by the decoder, predicting an individual detection set $\textbf{Y}_{g}$ for the input image. 
In total, this provides $G$ such individual yet \vth{overlapping} detection sets $\mathcal{Y}=\left\{ \textbf{Y}_{1},\textbf{Y}_{2},...,\textbf{Y}_{G} \right\}$ for the same image in a single forward pass.
The significant overlap between these $G$ detection sets is a key characteristic that unlocks ensemble-based uncertainty estimation.

\subsubsection{Clustering and Aggregation}
To form meaningful clusters among all the detections in $\mathcal{Y}$, we apply the Basic Sequential Algorithmic Scheme (BSAS)~\cite{b20} clustering method following existing works~\cite{b9, b23}.
Each cluster is composed of detections that share high mutual Intersection-over-Union (IoU) scores and have the same predicted class label.
For each cluster, we aggregate the detections within the cluster to obtain a final detection, along with the corresponding semantic and spatial uncertainty estimates. 

The following subsections will provide a detailed explanation of these key components.

\subsection{Additional Groups of Object Queries}
During inference, the additional groups of object queries must satisfy two key requirements: strong detection capability and diversity.
First, all query groups in $\mathcal{Q}$ must be able to accurately detect objects in the image.
Second, the object queries across different groups should be diverse. 
\vth{As} reliable ensemble-based uncertainty estimation relies on diverse individual predictions, the query groups must be diverse since they \vth{share the same model for detection.}
We have found that Group DETR~\cite{b18} fulfills both requirements, making it a seamless fit for our \name{} method.


As discussed in the Related Work section, Group DETR~\cite{b18} is a \vth{general} training method designed to accelerate the training of DETR-like models.
During training, Group DETR randomly initializes $G$ groups of learnable object queries $\mathcal{Q}=\left\{ \textbf{Q}_{1},\textbf{Q}_{2},...,\textbf{Q}_{G} \right\}$, where each query group $\textbf{Q}_{g} \in \mathcal{Q}$ contains $N$ queries. 
This random initialization introduces stochasticity across query groups.
Each query group $\textbf{Q}_{g}$ is then separately transformed by the decoder and processed by the task heads to predict a complete detection set  $\textbf{Y}_{g}$.
This results in a total of $G$ individual detection sets $\mathcal{Y}=\left\{ \textbf{Y}_{1},\textbf{Y}_{2},...,\textbf{Y}_{G} \right\}$.
The final loss for training is calculated as the average of the individual Hungarian loss between each detection set $\textbf{Y}_{g}$ and the same set of GT objects $\bar{\textbf{Y}}$.
As a result, each query group is learned separately without any inter-group interference, which reinforces the stochasticity among them. 
During inference, Group DETR only uses the first query group $\textbf{Q}_{1}$ to predict a single detection set $\textbf{Y}_{1}$, while the remaining query groups are discarded, despite their equivalent detection capabilities.
However, we propose that these additional query groups can be leveraged for uncertainty estimation, and also to ensemble results towards improved detection accuracy.

\begin{figure}[htbp]
\centerline{\includegraphics[width=1.0\columnwidth]{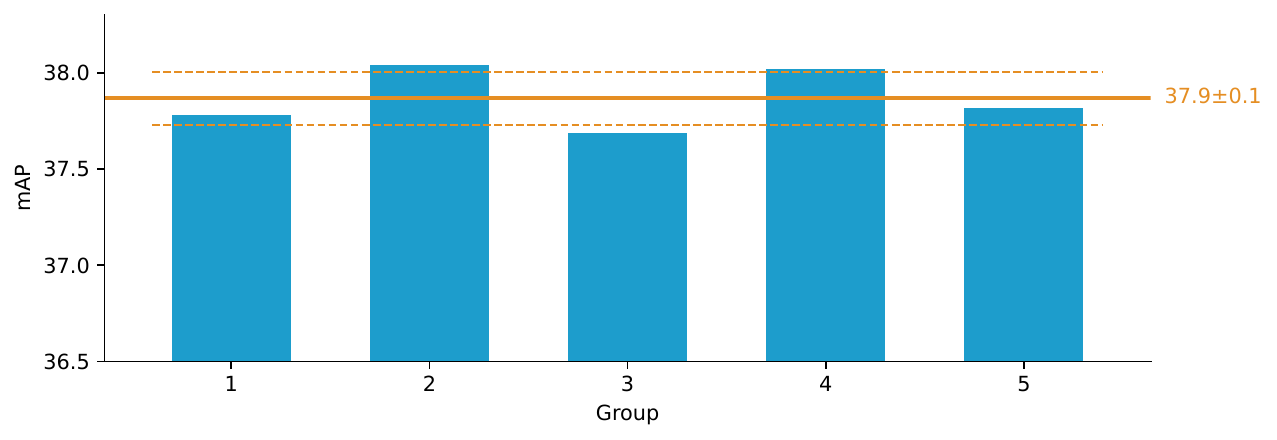}}
\vspace{-10pt}
\caption{
The detection performance per query group is similar.
The mAP scores are obtained by training a Conditional DETR~\cite{b15} with Group DETR~\cite{b18} on the Cityscapes dataset~\cite{b12}, using five query groups.
}
\label{fig:map}
\vspace{-2mm}
\end{figure}

Fig.~\ref{fig:map} demonstrates the strong detection capability of each query group, as measured by mean Average Precision (mAP).
\begin{figure}[htbp]
\centerline{\includegraphics[width=1.0\columnwidth]{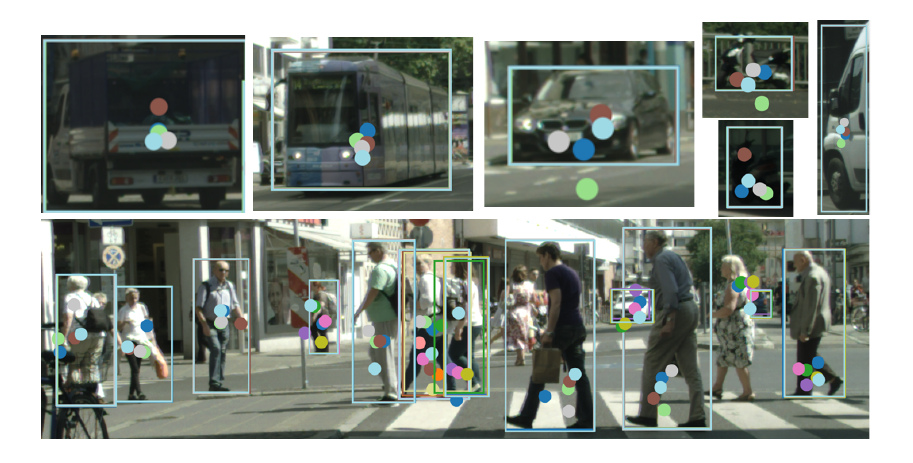}}
\vspace{-10pt}
\caption{
Visualizations of detection clusters and the corresponding object queries (represented as center reference points) from different query groups on the Cityscapes dataset~\cite{b12}.
Different colors are used to distinguish between groups.
The diversity of reference points detecting the same ground-truth objects is evident.
The reference points and bounding boxes may overlap.
}
\label{fig:refpoints}
\vspace{-4mm}
\end{figure}
We evaluate the detection performance of each of the $G=5$ query groups and observe that their performance is similar, indicating that all query groups are capable of accurately detecting objects.
This observation aligns with the findings reported in the Group DETR paper\cite{b18}.
To illustrate the diversity across these query groups, we visualize the detections of some exemplary clusters in Fig.~\ref{fig:refpoints}, along with the corresponding object queries (represented as reference points) from different groups.
We observe that the initial reference points, which ultimately lead to the detection of the same GT objects, are indeed diverse.
This observation verifies our hypothesis that stochasticity does exist across query groups, which is crucial for uncertainty estimation.

In summary, we show that Group DETR already meets the necessary requirements for reliable ensemble-based uncertainty estimation, making it an ideal choice as our baseline model. 
Furthermore, because Group DETR is a training method that can be applied to virtually any DETR-like model, our \name{} method inherits this flexibility.

\subsection{Attention Mask}
In contrast to the original Group DETR, \name{} does not discard any query groups during inference. 
Instead, we leverage all the high-quality, learned query groups to predict multiple, \vth{overlapping} detection sets  $\mathcal{Y}=\left\{ \textbf{Y}_{1},\textbf{Y}_{2},...,\textbf{Y}_{G} \right\}$ for the given image.
This redundancy is crucial for enabling ensemble-based uncertainty estimation. 
Recognizing that multiple highly correlated predictions can negatively impact the accuracy of uncertainty estimates~\cite{b6}, and drawing inspiration from DN-DETR~\cite{b17} \crv{and~\cite{b4}}, we design an attention mask for the decoder to ensure that each detection set in $\mathcal{Y}$ is predicted independently of the others during inference.

As mentioned earlier, the decoder's self-attention allows for interactions between queries, while the cross-attention enables each query to individually attend to the image feature $\textbf{X}$ without inter-query interference.
Therefore, to prevent interactions between queries of different groups, a self-attention mask is sufficient.
Formally, the transformer decoder receives a total of $W=G\cdot N$ queries (concatenated in the group order), where $G$ is the number of query groups and $N$ is the number of queries per group.
These queries, along with the image feature $\textbf{X}$, are fed as input to the decoder.
The self-attention mask, denoted as $M_\text{SA} = [m^\text{SA}_{ij}]_{W \times W}$, is defined as follows:
\begin{equation}
    m^\text{SA}_{ij} = 
    \begin{cases}
        1 & \text{if} \ \lfloor \frac{i}{N} \rfloor \neq \lfloor \frac{j}{N} \rfloor ;\\
        0 & \text{otherwise},
    \end{cases}
\end{equation}
where $m^\text{SA}_{ij}=1$ stands for that $i$-th cannot attend to $j$-th query and $m^\text{SA}_{ij}=0$ otherwise.
As visualized in Fig.~\ref{fig:overview}, this mask effectively blocks interactions between queries of different groups while fully enabling query interactions within the group. 
This allows the decoder to independently and simultaneously transform each query group, predicting $G$ individual detection sets (\ie, $\mathcal{Y}$) in \textit{a single forward pass}.
\vtwo{
In practice, we achieve the same functionality as $M_\text{SA} = [m^\text{SA}_{ij}]_{W \times W}$ by processing the query groups in the batch dimension. 
This avoids the $O((W^{2}))$ self-attention memory explosion, leading to greater efficiency.
}

\subsection{Clustering and Aggregation}
\subsubsection{Clustering}
Since each detection set $\textbf{Y}_{g} \in \mathcal{Y}$ is a complete set of detections for the same input image, there is significant overlap between the $G$ detection sets.
For example, a single GT instance can be independently detected multiple times by queries of different query groups.
To group these detections into meaningful clusters, we apply the Basic Sequential Algorithmic Scheme (BSAS)~\cite{b20} method following existing works~\cite{b9,b23}.
BSAS sequentially clusters detections based on two criteria: 1) the Intersection-over-Union (IoU) between bounding boxes and 2) the predicted class label.
For each detection, if its IoU with any existing cluster exceeds a predefined threshold $\theta$ and its predicted label matches the cluster's label, it is added to that cluster. 
Otherwise, the detection initiates a new cluster.
Consequently, each cluster comprises detections that share high IoU scores and the same predicted class labels.
We begin by sorting all detections in $\mathcal{Y}$ according to their confidence scores of class prediction, from highest to lowest, and then apply the BSAS method to form clusters,
\begin{equation}
\mathcal{C}=\left\{ \textbf{C}_{1},\textbf{C}_{2},...\textbf{C}_{M} \right\}=\text{BSAS}(\text{Sort}(\mathcal{Y})|\theta)
\label{bsas}
\end{equation}
where $\mathcal{C}$ is the set of clusters, $M$ is the total number of clusters.
\vth{Occasionally, we observed that the number of detections within a single cluster exceeded the number of detection sets $G$.
We identified two causes: (1) a single target being detected $\gt 1$ times by the same detection set, and (2) detections of closely located objects being clustered together. 
To mitigate the second cause, we attempted to restrict each cluster to contain $\le 1$ detection from any detection set.
However, in practice, this restriction led to slightly worse results due to an increase in false positives caused by (1).
Therefore we opted to use the standard BSAS method without this constraint in all experiments.}
The visualizations of some exemplary clusters are shown in Fig.~\ref{fig:refpoints}.

\subsubsection{Aggregation}
Since each cluster contains detections likely corresponding to the same GT instance, we aggregate the detections within each cluster to obtain a final detection and estimate the corresponding uncertainty.
For a given cluster $\textbf{C}_{m} \in \mathcal{C}$, we 
calculate its final confidence score $c_{m}$ using the equation below:
\begin{equation}
c_{m}=\alpha_{m}\cdot max\left( c_{i} \right) \;\;\;\;\;  \forall \textbf{y}_{i}=\left\{\textbf{b}_i, c_i, k_i \right\} \in \textbf{C}_m
    \label{eq:final_conf}
\end{equation}
where $\alpha_{m}=min( \left| \textbf{C}_m \right|,G) / G$, $\left| \textbf{C}_m \right|$ is the number of detections in the cluster, and $G$ is the number of query groups.
The intuition is that a cluster with a small number of detections may indicate that only a few query groups detected the object, while the others predicted it as background (\eg, a false positive).
Therefore, we use $\alpha_{m}$ to reduce the confidence score in such cases, while keeping the maximum score when all query groups detect the object, indicating high certainty. 
The final bounding box $\textbf{b}_{m}$ and the corresponding spatial uncertainty, measured by the variance $\Sigma_{m} \in \mathbb{R}^{4\times 4}$, are calculated as the following:
\begin{equation}
\textbf{b}_{m}=\sum_{i=1}^{\left| \textbf{C}_{m} \right|}\text{Softmax}\left( c_i \right)\cdot \textbf{b}_{i} 
\label{final_box}
\end{equation}
\vspace{-20pt}
\begin{equation}
\Sigma_{m} = \sum_{i=1}^{\left| \textbf{C}_{m} \right|} \text{Softmax}(c_i) \cdot (\textbf{b}_i - \textbf{b}_m)(\textbf{b}_i - \textbf{b}_m)^T
\label{final_box_var}
\vspace{-4mm}
\end{equation}
where $\text{Softmax}(c_i) = e^{c_i}/{\sum_{j=1}^{\left| \textbf{C}_m \right|}e^{c_j}}$.
Confidence-weighted averaging produces a more robust and accurate final bounding box by prioritizing high-confidence detections, which are \vth{likely} more precise, while down-weighting noisy, low-confidence outliers. 
Similarly, the weighted variance provides a more reliable spatial uncertainty estimate because it primarily reflects the disagreement among the most trustworthy predictions, preventing outliers from inflating the estimate.

Through this process, each cluster yields a final detection with uncertainty estimates, denoted as $\textbf{d}_m=\left\{ \textbf{b}_m, \Sigma_{m},k_m, c_m \right\}$, where the class label $k_m$ is the same as the cluster's label.
This results in a total of $M$ final detections for a given image, represented as $\textbf{D}=\left\{ \textbf{d}_1,\textbf{d}_2,...,\textbf{d}_M \right\}$.
Optionally, a confidence threshold can be applied to remove detections with low confidence scores.
Following the definition provided in~\cite{b25}, \vth{\name{} enables} \textit{Probabilistic Object Detection} because the outputs now include not only object detections but also estimates of both semantic and spatial uncertainty associated with those detections.
\crv{This offers potential safety assurance for safety-critical driving tasks like occupancy prediction~\cite{b21} and end-to-end autonomous driving~\cite{b11} by triggering early human intervention during uncertain events.}

\begin{table*}[t!]
  \small
  \centering
  \setlength{\tabcolsep}{1mm}
    \caption{\textbf{Comparison of \name{} with existing methods.} 
    $^*$measurements are reported for the transformer decoder only.
    $^\dagger$denotes our parallelized version. Best results in \textbf{bold}, second best \underline{underlined}, highest latency and parameters in \color{red}{red}.
    }
    \label{tab:sotacom}%
        \begin{tabular}{rc|c|c|c|c|cc}
\cmidrule{2-8}          & \multirow{2}[2]{*}{Dataset} & \multirow{2}[2]{*}{Method} & \textit{Uncertainty \& Acc.} & \textit{Calibration} & \textit{Accuracy} & \multicolumn{2}{c}{\textit{Efficiency}} \\
          &       &       & $\uparrow$PDQ(thr. 0.3) & $\downarrow$D-ECE(thr. 0.3) & $\uparrow$mAP & $\downarrow$Latency(ms)$^*$ & $\downarrow$\#Para(M) \\
\cmidrule{2-8}          & \multirow{5}[2]{*}{Cityscapes~\cite{b12}} & Deterministic & 9.4   & 11.8  & 37.8  & 10.7  & 43.17 \\
          &       & MC-Dropout$^\dagger$ & 18.9  & 10.9  & 38.5  & 18.4($\times 1.7$)  & 43.17 \\
          &       & Deep Ensembles & \underline{19.3}  & \textbf{9.8} & \underline{38.8}  & 53.4($\times \textcolor{red}{5.0}$)  & 89.27 (+\textcolor{red}{$107$}\%) \\
          &       & \cellcolor[rgb]{ .851,  .851,  .851}GroupEnsemble (ours) & \cellcolor[rgb]{ .851,  .851,  .851}18.8 & \cellcolor[rgb]{ .851,  .851,  .851}\underline{10.4} & \cellcolor[rgb]{ .851,  .851,  .851}38.7 & \cellcolor[rgb]{ .851,  .851,  .851}18.4($\times 1.7$) & \cellcolor[rgb]{ .851,  .851,  .851}43.48 (+0.7\%) \\
          &       & \cellcolor[rgb]{ .851,  .851,  .851}MC-GroupEnsemble (ours) & \cellcolor[rgb]{ .851,  .851,  .851}\textbf{21.4} & \cellcolor[rgb]{ .851,  .851,  .851}\underline{10.4} & \cellcolor[rgb]{ .851,  .851,  .851}\textbf{39.2} & \cellcolor[rgb]{ .851,  .851,  .851}18.4($\times 1.7$) & \cellcolor[rgb]{ .851,  .851,  .851}43.48 (+0.7\%) \\
\cmidrule{2-8}          & \multirow{5}[2]{*}{Foggy Cityscapes~\cite{b14}} & Deterministic & 9.5   & 17.1  & 26.1  & 10.7  & 43.17 \\
          &       & MC-Dropout$^\dagger$ & \underline{17.7}  & 14.6  & \underline{26.4}  & 18.4($\times 1.7$)  & 43.17 \\
          &       & Deep Ensembles & 17.5  & \textbf{12.1} & \underline{26.4}  & 53.4($\times \textcolor{red}{5.0}$)  & 89.27 (+\textcolor{red}{$107$}\%) \\
          &       & \cellcolor[rgb]{ .851,  .851,  .851}GroupEnsemble (ours) & \cellcolor[rgb]{ .851,  .851,  .851}17.2 & \cellcolor[rgb]{ .851,  .851,  .851}14.4 & \cellcolor[rgb]{ .851,  .851,  .851}26.2 & \cellcolor[rgb]{ .851,  .851,  .851}18.4($\times 1.7$) & \cellcolor[rgb]{ .851,  .851,  .851}43.48 (+0.7\%) \\
          &       & \cellcolor[rgb]{ .851,  .851,  .851}MC-GroupEnsemble (ours) & \cellcolor[rgb]{ .851,  .851,  .851}\textbf{19.1} & \cellcolor[rgb]{ .851,  .851,  .851}\underline{14.0} & \cellcolor[rgb]{ .851,  .851,  .851}\textbf{26.6} & \cellcolor[rgb]{ .851,  .851,  .851}18.4($\times 1.7$) & \cellcolor[rgb]{ .851,  .851,  .851}43.48 (+0.7\%) \\
\cmidrule{2-8}          & \multirow{5}[2]{*}{COCO~\cite{b13}} & Deterministic & 9.7   & 11.1  & 43.1  & 10.7 & 43.66 \\
          &       & MC-Dropout$^\dagger$ & \underline{17.4}  & 10.4  & 43.2  & 18.4($\times 1.7$)  & 43.66 \\
          &       & Deep Ensembles & \textbf{19.1} & \textbf{9.2} & \textbf{43.8} & 53.4($\times \textcolor{red}{5.0}$)  & 89.86 (+\textcolor{red}{$106$}\%)\\
          &       & \cellcolor[rgb]{ .851,  .851,  .851}GroupEnsemble (ours) & \cellcolor[rgb]{ .851,  .851,  .851}16.8 & \cellcolor[rgb]{ .851,  .851,  .851}\underline{10.3} & \cellcolor[rgb]{ .851,  .851,  .851}\underline{43.6} & \cellcolor[rgb]{ .851,  .851,  .851}18.4($\times 1.7$) & \cellcolor[rgb]{ .851,  .851,  .851}43.96 (+0.7\%)\\
          &       & \cellcolor[rgb]{ .851,  .851,  .851}MC-GroupEnsemble (ours) & \cellcolor[rgb]{ .851,  .851,  .851}\textbf{19.1} & \cellcolor[rgb]{ .851,  .851,  .851}\underline{10.3} & \cellcolor[rgb]{ .851,  .851,  .851}\underline{43.6} & \cellcolor[rgb]{ .851,  .851,  .851}18.4($\times 1.7$)& \cellcolor[rgb]{ .851,  .851,  .851}43.96 (+0.7\%) \\
\cmidrule{2-8}    \end{tabular}%
  \label{tab:addlabel}%
    \vspace{-2mm}
\end{table*}%
\section{EXPERIMENTS}
\subsection{Experiment Setup}
\subsubsection{\textbf{Metrics}}
To provide a comprehensive comparison between GroupEnsemble, MC-Dropout, and Deep Ensembles, we evaluate them across four categories:
\paragraph{Detection Accuracy}
We use the standard mean Average Precision (mAP)~\cite{b13}, which measures the average precision across all classes and multiple IoU thresholds to assess overall detection performance.
\paragraph{Uncertainty Quality}
We employ Probabilistic Detection Quality (PDQ)~\cite{b25} as our primary metric for uncertainty-aware performance. 
It assesses both detection accuracy and the quality of the predicted spatial and label uncertainties.

\paragraph{Calibration}
We measure Detection Expected Calibration Error (D-ECE)~\cite{b26} to quantify how well a model's predicted confidence scores align with its actual precision.

\paragraph{Computational Efficiency}
Our efficiency assessment is based on two metrics: latency, which quantifies the processing speed, and the number of parameters, which serves as an indicator of the model's size and memory footprint.

\subsubsection{\textbf{Dataset}}
Our experiments are conducted on three diverse datasets to cover both autonomous driving and general detection scenarios:
\paragraph{Cityscapes~\cite{b12}}
A large dataset of urban street scenes, using 2,975 images for training and 500 for validation.

\paragraph{Foggy Cityscapes~\cite{b14}}
A synthetic version of Cityscapes with simulated fog, used to test model robustness to domain shift.
The dataset features three visibility levels: 600m, 300m, and 150m, which correspond to attenuation coefficients of 0.005, 0.01, and 0.02.

\paragraph{COCO~\cite{b13}}
A widely-used, large-scale benchmark containing 80 object classes for general object detection.

\subsubsection{\textbf{Implementation Details}}
Our experimental baseline is Group DETR~\cite{b18}, built upon Conditional DETR~\cite{b15} with a ResNet-50~\cite{b27} backbone and five groups of object queries ($G=5$).
To prevent overfitting, dropout layers~\cite{b28} with a probability of 0.1 are applied throughout the model during training.
For the COCO dataset, we follow the official Group DETR training protocol. 
For Cityscapes, we fine-tune a COCO-pretrained model on 8 driving-related classes (person, rider, car, truck, bus, train, motorbike, and bicycle) at a higher image resolution.
\vth{Performance under} domain-shift is evaluated on the most challenging Foggy Cityscapes variant with 150m visibility.
All models are trained for 50 epochs on eight NVIDIA A100 GPUs, \vth{and the inference and latency measurement are conducted on one NVIDIA A6000 GPU.}

To isolate the effects of each ensembling technique, we used a shared, deterministic encoder and an identical clustering method (\ie, BSAS with $\theta$ \vth{empirically set to 0.7}) for all methods. 
The differences between GroupEnsemble, MC-Dropout, and Deep Ensembles were therefore implemented only in the transformer decoder and, where applicable, the task heads.
\crv{
The Deep Ensembles method uses five models with a frozen, pretrained encoder, but each model has independently trained decoders and task heads initialized with different random seeds.
To reduce memory consumption during inference with Deep Ensembles, we perform its multiple forward passes sequentially following~\cite{b29}.
}
\crv{
For MC-Dropout, we also leverage DETR's parallelism to speed up its inference. 
This is achieved by duplicating the first query group five times and passing the concatenated groups through the decoder in a single forward pass.}
We apply a self-attention mask, identical to the one used in GroupEnsemble, and activate all the decoder's dropout layers. 
The key to this parallel implementation is that the dropout layers generate statistically independent query masks over the different query groups, leading to diverse detection results between the groups.
\begin{figure*}[t!]
      \centering
      \includegraphics[width=1.0\textwidth]{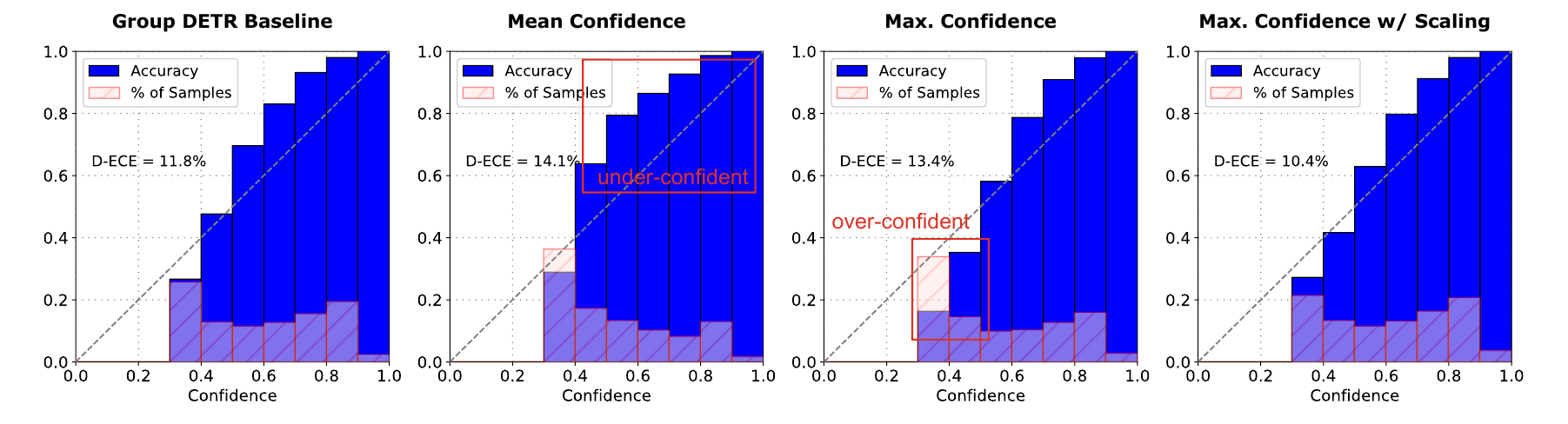}
      \vspace{-17pt}
      \caption{
        \textbf{D-ECE plots of the different confidence aggregation methods.}
        The calibration plots reveal that using the mean confidence score of clusters as the final scores results in under-confident detections (as highlighted in the 2nd plot), whereas using the maximum confidence without scaling leads to slightly over-confident detections (highlighted in the 3rd plot). 
        Both cases negatively impact calibration performance. 
        On the other hand, using the scaled maximum confidence, as shown in the 4th plot, effectively addresses these issues and improves calibration.
      }
      \label{fig:dece}
    \vspace{-5pt}
  \end{figure*}
  
\vth{We further combine our \name{} with MC-Dropout to create a hybrid approach, MC-GroupEnsemble.
A key difference is that MC-GroupEnsemble uses five different query groups to enhance stochasticity, rather than duplicating a fixed query group five times as in standard MC-Dropout.
}

\subsection{Experimental Results}
The evaluation results are summarized in Table~\ref{tab:sotacom}. 
While the deterministic Group DETR baselines show the lowest computational cost, their inability to estimate spatial uncertainty is reflected in their poor PDQ scores. 
At the other extreme, Deep Ensembles consistently provides the best calibration (lowest D-ECE) but is impractical for resource-limited applications like autonomous driving due to its massive parameter count (about 107\% more than baseline).
Our proposed method, GroupEnsemble, effectively bridges this gap. 
It achieves competitive performance to MC-Dropout across all datasets, outperforming the deterministic baselines in all metric categories. 
\crv{By parallelizing query group processing, both MC-Dropout and our GroupEnsemble are 66\% faster in processing time compared to their sequential counterparts, validating our design choice of leveraging the decoder's inherent parallelism to avoid the time-consuming multiple forward passes.}
\vth{Moreover, MC-GroupEnsemble demonstrated state-of-the-art results:} it achieved the best PDQ and mAP on both Cityscapes and the challenging Foggy Cityscapes domain-shift scenario, with only a negligible increase in model size (+0.7\%) over the standard MC-Dropout.
This suggests that GroupEnsemble and MC-Dropout are complementary, and their combination yields accurate uncertainty estimates at an acceptable computational cost.
\vtwo{Notably, MC-GroupEnsemble outperforms the strong Deep Ensembles in PDQ metrics, while requiring a fraction of the latency (66\% faster) and model parameters (51\% fewer). 
This highlights its great potential for real-time applications like autonomous driving.}

In summary, GroupEnsemble achieves a superior trade-off between performance and runtime and can be combined with MC-Dropout to further improve \crv{probabilistic} detection quality.
\vtwo{\crv{In the following}, we conduct ablation studies on Cityscapes using \name{} to validate our method's design choices.}

\subsection{Ablation Study}
\subsubsection{Number of Query Groups}
In this section, we analyze the impact of the number of query groups used by GroupEnsemble on the quality of the uncertainty estimates.
\begin{table}[t!]
    \centering
    \caption{\textbf{Ablation study on the number of query groups for GroupEnsemble.} 
    $^*$Measurements are reported for the transformer decoder only. Best results in \textbf{bold}.}
    \label{tab:num_g}%
    \setlength{\tabcolsep}{1mm}{%
    \begin{tabular}{c|ccc|cc}
    \toprule
    \#Groups & $\uparrow$PDQ & $\downarrow$D-ECE & $\uparrow$mAP & $\downarrow$Latency(ms)$^*$ & $\downarrow$\#Para(M) \\
    \midrule
    1     & 9.4   & 11.8  & 37.8  & 10.68 & 43.17 \\
    3     & 17.5  & 10.6  & 38.3  & 15.42$(\times1.4)$ & 43.32 \\
    \cellcolor[rgb]{ .851,  .851,  .851}\textbf{5}     & \cellcolor[rgb]{ .851,  .851,  .851}18.8  & \cellcolor[rgb]{ .851,  .851,  .851}10.4  & \cellcolor[rgb]{ .851,  .851,  .851}38.7  & \cellcolor[rgb]{ .851,  .851,  .851}18.39$(\times1.7)$ & \cellcolor[rgb]{ .851,  .851,  .851}43.48 \\
    7     & 19.4  & 10.4  & 38.7  & 22.51$(\times2.1)$ & 43.61 \\
    9     & \textbf{19.7}  & \textbf{10.3}  & \textbf{38.8}  & 26.55$(\times2.5)$ & 43.76 \\
    \bottomrule
    \end{tabular}%
      }
    \vspace{-1mm}
  \end{table}%

\begin{table}[t!]
    \centering
    \caption{\textbf{Ablation study on the aggregation methods for confidence scores.}  Best results in \textbf{bold}.}
    \label{tab:conf}%
    \setlength{\tabcolsep}{1mm}{%
    \begin{tabular}{c|ccc}
        \toprule
         Conf. Aggregation & $\uparrow$PDQ & $\downarrow$D-ECE & $\uparrow$mAP \\
        \midrule
        Baseline & 9.4   & 11.8  & 37.8 \\
        \midrule
        Mean Conf.  & 16.2  & 14.1  & 37.6 \\
        Max Conf.   & 15.9  & 13.4  & 38.4 \\
        \cellcolor[rgb]{ .851,  .851,  .851}\textbf{Max Conf. w/ Scaling} & \cellcolor[rgb]{ .851,  .851,  .851}\textbf{18.8} & \cellcolor[rgb]{ .851,  .851,  .851}\textbf{10.4} & \cellcolor[rgb]{ .851,  .851,  .851}\textbf{38.7} \\
        \bottomrule
        \end{tabular}%
      }
      \vspace{-3mm}
  \end{table}%
Table~\ref{tab:num_g} demonstrates that increasing the number of query groups used by GroupEnsemble results in higher PDQ scores, indicating more accurate uncertainty estimates. 
This is because more query groups generate more individual detection sets, leading to clusters containing more diverse detections of the same GT instance and, consequently, more accurate spatial uncertainty (variance) estimates.
Moreover, the transformer decoder's parallelism allows for simultaneous processing of all query groups without sequential repetition, resulting in relatively low latency even with a large number of query groups. 
This provides a significant speed advantage over sequential approaches like Deep Ensembles, which suffers from linearly scaled processing time w.r.t. the number of forward passes. 
\vtwo{We use five query groups by default for its balanced latency-performance trade-off.}

\subsubsection{Aggregation Strategy}
As defined in Eq.~\ref{eq:final_conf}, the final confidence score of a cluster is calculated by scaling its maximum confidence by the ratio of the number of detections to the number of query groups. 
We evaluate different confidence aggregation methods in Table~\ref{tab:conf} 
\vtwo{which differ in how they calculate final confidence scores for clusters with varying numbers of detections.}
\vtwo{Using the mean confidence performs worse than the deterministic Group DETR baseline across all three metrics. }
Using the maximum confidence without scaling improves PDQ and mAP but worsens the calibration metric D-ECE. 
On the other hand, applying the scaling method to the maximum confidence achieves improved results compared to the baseline. 
To better understand the underlying reasons for these results, we present calibration plots of D-ECE for the different aggregation methods in Fig.~\ref{fig:dece}.
These plots show the relationship between predicted confidence and actual detection accuracy (precision). 
Ideally, the plots should align with the diagonal, indicating good calibration. 
The baseline (1st plot) is already under-confident, as the actual accuracy is higher than the predicted confidence. 
\vtwo{Using the mean confidence (2nd plot) further exacerbates this under-confidence.}
Conversely, using the maximum confidence without scaling leads to over-confidence in certain regions (highlighted in red in the 3rd plot), resulting in poor calibration scores. 
Finally, scaling the maximum confidence (4th plot) improves calibration compared to the baseline,
\vtwo{as the scaling incorporates implicit uncertainty information from the cluster size, reducing confidence for clusters with few detections (potential false positives).}

\section{CONCLUSION}
\vtwo{This paper proposes \name{}, an efficient and effective uncertainty estimation method for DETR-based object detection. 
\name{} leverages the inherent parallelism of the DETR decoder to achieve single-pass uncertainty estimation without sequential repetition. 
Results on three datasets show \name{} performs comparably to MC-Dropout, and the hybrid MC-GroupEnsemble outperforms Deep Ensembles on certain metrics with significantly lower cost, making it promising for online autonomous driving.
}


\end{document}